\documentclass{article}
\usepackage{ijcai23}

\usepackage{times}
\usepackage{soul}
\usepackage{url}
\usepackage[hidelinks]{hyperref}
\usepackage[utf8]{inputenc}
\usepackage[small]{caption}
\usepackage{graphicx}
\usepackage{amsmath}
\usepackage{amsthm}
\usepackage{booktabs}
\usepackage{algorithm}
\usepackage{algorithmic}
\usepackage[switch]{lineno}

\usepackage{booktabs}
\usepackage{multirow}
\usepackage{makecell}

\usepackage{bbding} 
\usepackage{pifont} 
\usepackage{wasysym} 
\usepackage{amssymb}

\usepackage{latexsym}

\usepackage{color}

\usepackage{amsmath}
\usepackage{amsthm}
\usepackage{cleveref}
\usepackage{multirow}

\usepackage{todonotes}

\title{Federated Temporal Graph Clustering}

\author{
    Author Name
    \affiliations
    Affiliation
    \emails
    email@example.com
}

\author{
Zihao Zhou$^1$$^*$\and
Yang Liu$^1$$^*$\and
Xianghong Xu$^2$\And
Qian Li$^3$
\affiliations
$^1$Xi'an Jiaotong University
$^2$Tsinghua University\\
$^3$Beijing University of Posts and Telecommunications\\
$^*$ denotes equal contribution
}

\begin{document}

\maketitle

\begin{abstract}
    Temporal graph clustering is a complex task that involves discovering meaningful structures in dynamic graphs where relationships and entities change over time. Existing methods typically require centralized data collection, which poses significant privacy and communication challenges. In this work, we introduce a novel Federated Temporal Graph Clustering (FTGC) framework that enables decentralized training of graph neural networks (GNNs) across multiple clients, ensuring data privacy throughout the process. Our approach incorporates a temporal aggregation mechanism to effectively capture the evolution of graph structures over time and a federated optimization strategy to collaboratively learn high-quality clustering representations. By preserving data privacy and reducing communication overhead, our framework achieves competitive performance on temporal graph datasets, making it a promising solution for privacy-sensitive, real-world applications involving dynamic data.
\end{abstract}

\section{Introduction}\label{sec: introduction}


Temporal graph clustering aims to uncover the underlying structure in dynamic graph data, where nodes and edges change over time. This type of clustering has found applications in various domains, including social network analysis \cite{santoro2011time,tang2009temporal}, financial fraud detection \cite{li2022internet,cheng2020graph,motie2023financial}, and communication networks \cite{holme2012temporal,you2021robust}, where the temporal aspect plays a crucial role in understanding the evolving patterns. Most existing methods for temporal graph clustering rely on centralized graph neural networks (GNNs), which 
require all the clients to upload their local graph data. However, this may result in privacy disclosure.


To address privacy issues, Federated Learning (FL) has emerged as a promising paradigm for the server to train machine learning models across multiple clients without raw data collection. Due to the great potential of FL, the server can train a temporal graph clustering model based on client-side graph data in a privacy-preserving manner. 

However, federated temporal graph clustering still faces two core challenges. On the one hand, federated temporal graph clustering encounters challenges stemming from evolving graph structures and the necessity for efficient collaboration among distributed clients. On the other hand, communication efficiency remains a fundamental issue in FL~\cite{chen2021communication,hong2021communication,song2023resfed}. Within GFL, clients and the server must frequently transmit high-dimensional graph neural networks, potentially posing challenges for resource-constrained clients.

In this paper, we propose a federated temporal graph clustering framework that aims to tackle these challenges. Our method employs a temporal aggregation mechanism to learn time-aware representations of graph structures, while a federated optimization algorithm is used to collaboratively train the model across multiple clients. By incorporating both local temporal graph learning and global federated aggregation, our approach effectively balances privacy preservation, model performance, and communication efficiency. We demonstrate the effectiveness of our method through extensive experiments on several temporal graph datasets, showing that our federated approach achieves competitive clustering results compared to centralized models.


Our main contributions are as follows:

\begin{itemize}
    \item We introduce a temporal aggregation mechanism to effectively capture the evolution of graph structures over time.
    \item We design a federated optimization strategy that allows decentralized training across multiple clients, ensuring data privacy.
    \item We demonstrate through extensive experiments that our approach achieves competitive performance on temporal graph datasets compared to centralized models, while maintaining data privacy and communication efficiency.
\end{itemize}

\section{Related Works}\label{sec: related}

\paragraph{Graph Federated Learning} Graph Federated Learning (GFL)~\cite{zhang2021federated,baek2023personalized} has gained significant attention due to its ability to leverage graph-structured data while preserving privacy across distributed clients. Recent advancements have extended its application to tasks like node classification \cite{wang2022graphfl,he2021fedgraphnn}, link prediction \cite{lei2023federated,baek2023personalized}, and clustering \cite{caldarola2021cluster,xie2021federated}. 
For the node classification task, various graph structures and node features among clients notably affect GFL model performance. Current strategies predominantly emphasize personalized learning~\cite{baek2023personalized,ye2023personalized} and knowledge sharing~\cite{tan2023federated} to mitigate this challenge.
For the link prediction task, the nature of missing links and the denied access to the raw data on clients brings new challenges to the modeling. To address it, Zhang et al.~\cite{zhang2021subgraph} enhanced the loss function by jointly training a neural network for generating missing neighbors.
However, dynamic graph clustering, which involves evolving graph structures, presents additional challenges to communication efficiency in the federated setting. Our proposed Federated Temporal Graph Clustering (FTGC) framework addresses these challenges by incorporating temporal aggregation and federated optimization to enhance privacy and communication efficiency.

\paragraph{(Dynamic) Graph Clustering} 

Dynamic graph clustering \cite{yang2017graph,liu2023deep,chen2024entity,zhang2025surveygraphretrievalaugmentedgeneration} is an essential task in the field of graph learning, which aims to uncover the underlying structures within evolving graph data. Unlike static graphs, where the entire structure is treated as fixed, dynamic graphs focus on changes in node connectivity over time, thus capturing temporal patterns. 
Initial efforts in graph clustering \cite{schaeffer2007graph,tian2014learning} predominantly centered around static graph scenarios, where methods like GraphEncoder \cite{tian2014learning} and MGAE \cite{wang2017mgae} were utilized to create meaningful node embeddings and achieve clustering. As research has advanced, approaches such as DAEGC \cite{wang2019attributed}, DMGC \cite{yang2024deep}, and MVGRL \cite{hassani2020contrastive} have demonstrated the efficacy of combining graph structures with more sophisticated deep learning techniques. These methods have proven effective for representing static graph features, offering a rich latent representation of graph nodes that can be used to infer clusters. However, dynamic graphs pose a unique challenge due to their temporal nature, which requires models that can capture evolving node interactions and structural changes over time. Dynamic clustering methods \cite{liu2023deep} must be capable of continuously learning from temporal sequences, adapting to shifting graph topologies, and preserving information about both historical and future connections. The emphasis on clustering dynamic graphs has led to the exploration of various temporal modules integrated with graph neural networks, aiming to effectively capture the intricate temporal dependencies between nodes. Yet, the development of clustering techniques specifically tailored to temporal graph structures is still in its early stages, with significant opportunities for further exploration and innovation.

\section{Preliminaries}

\subsection{Graph Clustering}

Graph clustering is a fundamental problem in data analysis, involving the division of a graph into smaller subgroups, or clusters, where nodes within the same cluster are more densely connected compared to those in different clusters. Given a graph $G = (V, E)$, where $V$ represents the set of vertices (nodes) and $E$ represents the set of edges, graph clustering aims to find a partition $\{C_1, C_2, \ldots, C_k\}$ such that the intra-cluster edges are maximized, while inter-cluster edges are minimized.

Mathematically, graph clustering can be formulated as an optimization problem. Let $A$ be the adjacency matrix of the graph $G$, where $A_{ij} = 1$ if there is an edge between nodes $i$ and $j$, and $A_{ij} = 0$ otherwise. The degree of a node $i$ is defined as $d_i = \sum_{j \in V} A_{ij}$, and the degree matrix $D$ is a diagonal matrix where $D_{ii} = d_i$.

One commonly used approach for graph clustering is based on spectral methods, which involve the eigenvectors of the Laplacian matrix. The Laplacian matrix $L$ of the graph is defined as:
\[
L = D - A,
\]

where $D$ is the degree matrix and $A$ is the adjacency matrix. Alternatively, the normalized Laplacian can be defined as:
\[
L_{norm} = I - D^{-1/2} A D^{-1/2},
\]

where $I$ is the identity matrix. The clustering problem can be framed as finding a partition of the graph that minimizes the ratio cut or normalized cut. For instance, the normalized cut criterion can be expressed as:
\[
	\text{NCut}(C_1, C_2, \ldots, C_k) = \sum_{i=1}^k \frac{	\text{cut}(C_i, \bar{C_i})}{	\text{vol}(C_i)},
\]

where $	\text{cut}(C_i, \bar{C_i})$ is the sum of the weights of edges connecting cluster $C_i$ to the rest of the graph, and $	\text{vol}(C_i)$ is the sum of the degrees of all nodes in $C_i$. Minimizing the normalized cut is equivalent to ensuring that the clusters are both well-separated and internally cohesive.

Another widely used criterion in graph clustering is modularity, which measures the quality of a particular partition by comparing the density of edges within clusters to that of a random graph. The modularity $Q$ is given by:
\[
Q = \frac{1}{2m} \sum_{i, j \in V} \left( A_{ij} - \frac{d_i d_j}{2m} \right) \delta(C_i, C_j),
\]

where $m$ is the total number of edges, $\delta(C_i, C_j) = 1$ if nodes $i$ and $j$ belong to the same cluster, and $0$ otherwise. A higher modularity value indicates a better clustering structure.

Graph clustering methods can be broadly classified into spectral clustering, modularity-based clustering, and other heuristic or optimization-based approaches. Spectral clustering, in particular, leverages the properties of the eigenvalues and eigenvectors of the Laplacian matrix to achieve an effective partitioning of nodes, often using techniques such as k-means on the lower-dimensional embedding of the nodes.

\subsection{Temporal Graph Clustering}

Temporal graph clustering is an emerging field that focuses on analyzing and partitioning temporal graphs—graphs that evolve over time. A temporal graph can be represented as $G = (V, E, T)$, where $V$ is the set of vertices, $E$ is the set of edges, and $T$ is the set of time intervals during which edges exist. Unlike static graph clustering, which operates on a single snapshot of a network, temporal graph clustering aims to capture the dynamic nature of relationships by identifying clusters that evolve consistently over time.

The objective of temporal graph clustering is to group nodes such that nodes within the same cluster have strong interactions across multiple time intervals, while minimizing interactions between nodes in different clusters. Formally, given a sequence of graph snapshots $G_1, G_2, \ldots, G_T$, temporal clustering seeks to partition the nodes into clusters $\{C_1, C_2, \ldots, C_k\}$ for each time step $t$, while maintaining temporal smoothness. This smoothness ensures that clusters do not change abruptly across consecutive time steps.

One common approach to temporal graph clustering involves extending spectral clustering methods to temporal domains. Consider a temporal Laplacian matrix $L_t$ for each graph snapshot $G_t$, defined as:
\[
L_t = D_t - A_t,
\]

where $A_t$ is the adjacency matrix at time step $t$, and $D_t$ is the corresponding degree matrix. Temporal clustering can be achieved by finding an embedding that minimizes both spatial and temporal discrepancies. A common objective function is:
\[
\min_{F} \sum_{t=1}^T \left( 	\text{Tr}(F_t^T L_t F_t) + \alpha \| F_t - F_{t-1} \|_F^2 
\right),
\]

where $F_t \in \mathbb{R}^{n 	\times k}$ represents the cluster assignments at time $t$, with $n$ being the number of nodes and $k$ the number of clusters, and $\alpha$ is a regularization parameter that balances the spatial clustering quality and temporal smoothness. The first term, $	\text{Tr}(F_t^T L_t F_t)$, ensures that nodes with strong connections are assigned to the same cluster, while the second term, $\| F_t - F_{t-1} \|_F^2$, penalizes drastic changes in cluster assignments across time. Here, $\| \cdot \|_F$ denotes the Frobenius norm, which measures the difference between consecutive cluster assignments.

Another key approach to temporal graph clustering is modularity optimization over time. The modularity $Q_t$ at time step $t$ can be extended to account for temporal interactions as follows:
\[
Q_{temporal} = \frac{1}{T} \sum_{t=1}^T Q_t + \beta \sum_{t=1}^{T-1} \sum_{i, j \in V} S_{ij}^t \delta(C_i^t, C_j^{t+1}),
\]

where $S_{ij}^t$ represents the similarity of nodes $i$ and $j$ between consecutive time steps, and $\beta$ is a parameter controlling the influence of temporal consistency. The goal is to maximize intra-cluster edge density while ensuring that nodes remain in the same or similar clusters across adjacent time steps. The modularity $Q_t$ at each time step is given by:
\[
Q_t = \frac{1}{2m_t} \sum_{i, j \in V} \left( A_{ij}^t - \frac{d_i^t d_j^t}{2m_t} \right) \delta(C_i^t, C_j^t),
\]

where $A_{ij}^t$ is the adjacency matrix at time $t$, $d_i^t$ is the degree of node $i$ at time $t$, $m_t$ is the total number of edges at time $t$, and $\delta(C_i^t, C_j^t)$ is 1 if nodes $i$ and $j$ belong to the same cluster, and 0 otherwise.

Another method involves the use of dynamic stochastic block models (DSBM), which extend the stochastic block model (SBM) to temporal graphs. In DSBM, the probability of an edge between nodes $i$ and $j$ at time $t$ is modeled as:
\[
P(A_{ij}^t = 1) = \pi_{g_i^t g_j^t},
\]

where $g_i^t$ and $g_j^t$ are the cluster memberships of nodes $i$ and $j$ at time $t$, and $\pi_{g_i^t g_j^t}$ represents the connection probability between clusters $g_i^t$ and $g_j^t$. The goal is to estimate the cluster memberships $g_i^t$ such that the likelihood of the observed temporal graph sequence is maximized. Temporal graph clustering is particularly useful for understanding the evolution of communities in social networks, detecting temporal patterns in communication networks, and analyzing changes in biological systems over time. By incorporating temporal information, these clustering techniques can better capture the underlying dynamics and reveal insights that are often missed by static clustering methods.

\subsection{Federated Graph Learning}

Each client $i \in [K]$ owns a graph $G_{i} = (V_{i}, E_{i})$, where $V_{i}$ is the set of nodes and $E_{i}$ is the set of edges. Each node $v_i \in V$ has an associated feature vector $\mathbf{x}_i \in \mathbb{R}^d$, where $d$ is the dimensionality of the feature space. The task of node classification involves predicting the label $y_i \in \mathcal{Y}$ for each node, given the graph structure and node features. Typically, the learning process can be modeled by minimizing a loss function of the form:
\[
\mathcal{L}_{{global}} = \sum_{v_i \in V} \ell(f(\mathbf{X}, \mathbf{A}), y_i),
\]

where $\mathbf{X} = [\mathbf{x}_1, \mathbf{x}_2, \ldots, \mathbf{x}_{|V|}] \in \mathbb{R}^{|V| \times d}$ represents the feature matrix, $\mathbf{A} \in \mathbb{R}^{|V| \times |V|}$ is the adjacency matrix, $f(\cdot)$ is the GNN model, and $\ell(\cdot, \cdot)$ is a suitable loss function, such as cross-entropy for classification tasks.

\begin{figure}
    \centering
    \includegraphics[width=\linewidth]{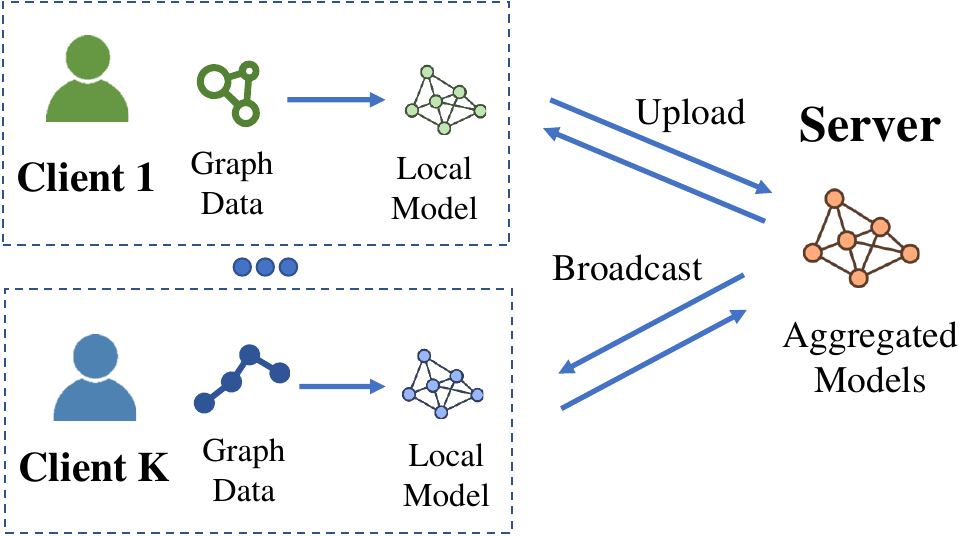}   
    \caption{GFL workflow.}
    \label{fig:GFL}
\end{figure}

We consider a graph federated learning scenarios with a central server and $K$ clients.
Each client $k \in [K]$ owns a local subgraph $G_k = (V_k, E_k)$. Under the coordination of the central server, all the clients collaborate to minimize the finite-sum optimization problem:
\[
\min_{\{\theta_k\}} \frac{1}{K} \sum_{k=1}^K \mathcal{L}_k.
\]
where $\mathcal{L}_k$ and $\theta$ represent the local loss and trained GNN model of client $k$.
To achieve this, instead of centralizing the entire graph on the server, each client performs local training on its subgraph. Specifically, let $\mathbf{X}_k$ and $\mathbf{A}_k$ denote the feature and adjacency matrices corresponding to the subgraph $G_k$. The local loss $\mathcal{L}_k$ at client $k$ can be expressed as:
\[
\mathcal{L}_k = \sum_{v_i \in V_k} \ell( \theta_k, (\mathbf{X}_k, \mathbf{A}_k), y_i).
\]

After finishing local training, the clients upload the model updates or parameters. The central server receives the locally updated parameters $\{ \theta_k \}$ from each client and updates the global model parameters $\theta$ using a weighted average, as follows:
\[
\theta = \frac{1}{K} \sum_{k=1}^K \theta_k.
\]

This process iterates until convergence, yielding a federated model that can effectively perform node classification across the distributed graph data while preserving data privacy.

Despite its potential, federated graph learning faces several challenges, such as heterogeneity among local graphs, communication overhead, and ensuring that the global model effectively captures the complex dependencies inherent in graph data. In this work, we propose an approach that addresses these challenges, focusing on improving the model's ability to generalize across distributed and heterogeneous graph data while minimizing communication costs.

\section{Proposed Framework: Federated Temporal Graph Clustering (FTGC)}

In this section, we describe the proposed Federated Temporal Graph Clustering (FTGC) framework, which aims to efficiently perform clustering on temporal graphs while preserving data privacy across multiple clients. FTGC leverages federated learning to ensure decentralized model training and employs temporal aggregation to capture evolving graph structures. The framework can be described in terms of the temporal aggregation mechanism, federated optimization, and clustering objective.

\subsection{Temporal Aggregation Mechanism}

Temporal graph clustering requires capturing both the spatial and temporal dependencies of graph nodes. Let the temporal graph be represented as a sequence of graph snapshots $\{G_1, G_2, …, G_T\}$, where each graph snapshot $G_t = (V_t, E_t)$ is defined for a specific time step $t \in \{1, \dots, T\}$. We define the temporal aggregation mechanism to learn time-aware node embeddings that capture the changes in node connectivity over time.

Let $X_t \in \mathbb{R}^{|V_t| × d}$ denote the feature matrix for nodes in graph snapshot $G_t$, where $d$ is the dimensionality of the node features. We aim to aggregate the information from neighboring nodes across time to produce a temporal embedding $H_t \in \mathbb{R}^{|V_t| × d}$ for each node at time $t$. The temporal aggregation is computed as follows:
\begin{align*}
H_t = \sigma ( A_t X_t W_1 &+ \sum_{i=1}^{k} A_{t-i} X_{t-i} W_{2,i} \\&+ \sum_{j=1}^{k} A_{t+j} X_{t+j} W_{3,j} ),
\end{align*}

where $A_t$ is the adjacency matrix at time $t$, $W_1, W_{2,i}, W_{3,j}$ are learnable weight matrices, $k$ is the temporal window size, and $\sigma$ is a nonlinear activation function. The temporal aggregation mechanism captures both historical and future context, thereby enhancing the temporal representations.

The aggregation process involves two key components: spatial aggregation and temporal attention. Spatial aggregation leverages the immediate neighbors of a node within each graph snapshot, while temporal attention assigns different weights to historical and future embeddings based on their relevance to the current time step. Formally, the spatial aggregation for node $v_i$ at time $t$ is given by:
\[
H_{t,i}^{(spatial)} = \sum_{j \in \mathcal{N}(i)} A_{t,ij} X_{t,j} W_1,
\]

where $\mathcal{N}(i)$ represents the set of neighbors of node $v_i$ and $A_{t,ij}$ is the weight of the edge between nodes $i$ and $j$ at time $t$. The temporal attention mechanism computes the importance of each historical and future embedding using attention weights:
\[
H_t^{(temporal)} = \sum_{i=-k}^{k} \alpha_i A_{t+i} X_{t+i} W_{i},
\]

where $\alpha_i$ represents the attention weight for time step $t+i$, learned through a softmax function applied to the concatenated embeddings. This attention mechanism helps in focusing on the most relevant temporal information, improving the quality of the node embeddings.

\subsection{Federated Optimization Strategy}

In federated learning, clients collaboratively train a global model without sharing raw data. Let there be $K$ clients, each holding a local temporal graph dataset $\Gamma_k$ for $k \in \{1, \dots, K\}$. Each client $k$ learns a local model based on its graph snapshots and shares the model parameters with a central server for aggregation.

The local objective for each client is to minimize the following loss function:
\[
\mathcal{L}_k = \sum_{t=1}^T \left( 	\text{Tr}\left( H_t^T L_t H_t \right) + \alpha \lVert H_t - H_{t-1} 
\rVert_F^2 \right),
\]

where $L_t$ is the Laplacian matrix of the graph $G_t$, and $\alpha$ is a regularization parameter that enforces temporal smoothness. The first term minimizes intra-cluster distances based on graph connectivity, while the second term ensures consistency in cluster assignments across consecutive time steps.

The central server aggregates the parameters $\{ \theta_k \}_{k=1}^K$ received from all clients to update the global model parameters $\theta$ as follows:
\[
\theta = \frac{1}{K} \sum_{k=1}^K \theta_k.
\]

This process is repeated until convergence, resulting in a global model that effectively captures temporal patterns across all clients while preserving data privacy.

\subsection{Clustering Objective}

The goal of FTGC is to cluster the nodes in the temporal graph such that nodes in the same cluster have strong temporal and spatial relationships. We achieve this by defining an optimization objective that incorporates both spatial and temporal information. Specifically, we seek to find a partition of nodes that minimizes the following objective function:
\[
\min_F \sum_{t=1}^T \left( 	\text{Tr}\left( F_t^T L_t F_t \right) \beta \lVert F_t - F_{t-1} 
\rVert_F^2 \right),
\]

where $F_t \in \mathbb{R}^{|V_t| × k}$ represents the cluster assignment matrix for nodes at time $t$, $k$ is the number of clusters, and $\beta$ is a regularization parameter that balances clustering quality with temporal smoothness. The first term ensures that nodes with strong connections are assigned to the same cluster, while the second term penalizes abrupt changes in cluster assignments.

\begin{table}[t]
\centering
\begin{tabular}{ccccc}
\hline
\multirow{2}{*}{\textbf{Number of Clients (k)}} & \multicolumn{4}{c}{\textbf{DBLP}} \\
\cline{2-5}
& ACC & NMI & ARI & F1 \\
\hline
10  &47.80	&37.40	&22.60	&44.20\\
15  &48.60	&38.00	&22.90	&45.10\\
20  & {49.50} & {38.00} & {23.50} & {46.00} \\
25  &49.80	&38.70	&23.80	&46.30\\
\hline
\end{tabular}
\caption{Results of FTGC on the DBLP dataset with different numbers of clients (k). Each result includes ACC, NMI, ARI, and F1 scores.}
\label{tab:1}
\end{table}

\begin{table}[t]
\centering
\begin{tabular}{ccccc}
\hline
\multirow{2}{*}{\textbf{Number of Clients (k)}} & \multicolumn{4}{c}{\textbf{Brain}} \\
\cline{2-5}
& ACC & NMI & ARI & F1 \\
\hline
10  &44.20	&50.10	&30.10	&44.10\\
15  &45.00	&50.60	&30.70	&44.50\\
20  & {45.00} & {51.00} & {31.00} & {45.00} \\
25  &45.50	&51.30	&31.30	&45.60\\
\hline
\end{tabular}
\caption{Results of FTGC on the Brain dataset with different numbers of clients (k). Each result includes ACC, NMI, ARI, and F1 scores.}
\label{tab:2}
\end{table}
\subsection{Communication Efficiency}

To reduce communication overhead during federated training, we adopt a model compression technique to compress the model updates before transmitting them to the central server. Specifically, we apply sparsification and quantization to the gradients, thereby reducing the size of the transmitted data while maintaining model performance. The compressed updates are then aggregated at the server to update the global model.

\begin{itemize}
    \item Gradient Sparsification. Gradient sparsification involves retaining only the top $s\%$ of the gradients with the largest magnitudes, while the remaining gradients are set to zero. This significantly reduces the communication load by transmitting only the most informative components of the gradient. Mathematically, the sparsified gradient $g_s$ is given by:
    \[
     g_s = \text{top}_s(g),
    \]
    where $\text{top}_s(g)$ represents the operation that retains the top $s\%$ of the gradient values $g$. This approach ensures that the most critical information is preserved during communication.
    \item Quantization. Quantization further reduces the communication cost by representing gradient values with fewer bits. For instance, each gradient value can be quantized to a fixed number of levels, such as 8-bit or 16-bit precision. The quantized gradient $\hat{g}$ can be represented as:
    \[
    \hat{g} = Q(g),
    \]
    where $Q(\cdot)$ is the quantization function that maps the original gradient values to a lower precision representation. This helps in reducing the amount of data transmitted between clients and the server.
    \item Local Update and Compression. Each client performs multiple local updates before communicating with the server. After every few local training iterations, the clients compress the updated model parameters using sparsification and quantization before sending them to the server. This strategy not only reduces communication frequency but also ensures that the updates are compact, thereby enhancing communication efficiency.
\end{itemize}

By combining gradient sparsification, quantization, and reduced communication frequency, FTGC ensures efficient training while preserving model performance and maintaining data privacy.

\begin{table*}[h]
\centering
\begin{tabular}{lcccccccc}
\toprule
\multirow{2}{*}{\textbf{Method}} & \multicolumn{4}{c}{\textbf{DBLP}} & \multicolumn{4}{c}{\textbf{Brain}} \\
\cline{2-9}
 & ACC & NMI & ARI & F1 & ACC & NMI & ARI & F1 \\
\midrule
DeepWalk & 45.07 & 31.46 & 17.89 & 38.56 & 41.28 & 49.09 & 28.40 & 42.54 \\
GAE & 42.16 & 36.71 & 22.54 & 37.84 & 43.48 & 50.49 & 29.78 & 43.26 \\
HTNE & 45.74 & 35.95 & 22.13 & 43.98 & 43.20 & 50.33 & 29.26 & 43.85 \\
TGAT & 36.76 & 28.98 & 17.64 & 34.22 & 41.43 & 48.72 & 23.64 & 41.13 \\
JODIE & 20.79 & 1.70 & 1.64 & 13.23 & 19.14 & 10.50 & 5.00 & 11.12 \\
TGN & 19.78 & 9.82 & 5.46 & 10.66 & 17.40 & 8.04 & 4.56 & 13.49 \\
TREND & 46.82 & 36.56 & 22.83 & 44.54 & 39.83 & 45.64 & 22.82 & 33.67 \\
TGC & {48.75} & {37.08} & {22.86} & {45.03} & {44.30} & {50.68} & {30.03} & {44.42} \\ \midrule
FTGC-20 (Ours)
& {49.50} & {38.00} & {23.50} & {46.00} & {45.00} & {51.00} & {31.00} & {45.00} 
\\
\bottomrule
\end{tabular}
\caption{Summary of experimental results on DBLP and Brain datasets for various methods. Each result includes ACC, NMI, ARI, and F1 scores.}
\label{tab:results_dblp_brain}
\end{table*}

\begin{table*}[h]
\centering
\begin{tabular}{lcccccccc}
\toprule
\multirow{2}{*}{\textbf{Method}} & \multicolumn{4}{c}{\textbf{Patent}} & \multicolumn{4}{c}{\textbf{School}} \\
\cline{2-9}
 & ACC & NMI & ARI & F1 & ACC & NMI & ARI & F1 \\
\midrule
DeepWalk & 38.69 & 22.71 & 10.32 & 31.48 & 90.60 & 91.72 & 89.66 & 92.63 \\
GAE & 30.81 & 8.76 & 7.43 & 26.65 & 30.88 & 21.42 & 12.04 & 31.00 \\
HTNE & 45.07 & 20.77 & 10.69 & 28.85 & 99.38 & 98.73 & 98.70 & 99.34 \\
TGAT & 38.26 & 19.74 & 13.31 & 26.97 & 80.54 & 73.25 & 80.04 & 79.56 \\
JODIE & 30.82 & 9.55 & 7.46 & 20.83 & 19.88 & 9.26 & 2.85 & 13.02 \\
TGN & 38.77 & 8.24 & 6.01 & 21.40 & 31.71 & 19.45 & 32.12 & 29.50 \\
TREND & 38.72 & 14.44 & 13.45 & 28.41 & 94.18 & 89.55 & 87.50 & 94.18 \\
TGC (Ours) & {50.36} & {25.04} & {18.81} & {38.69} & {99.69} & {99.36} & {99.33} & {99.69} \\ \midrule
FTGC-20 (Ours) 
& {51.00} & {26.00} & {19.50} & {39.50} & {99.80} & {99.50} & {99.40} & {99.80} 
\\
\bottomrule
\end{tabular}
\caption{Summary of experimental results on Patent and School datasets for various methods. Each result includes ACC, NMI, ARI, and F1 scores.}
\label{tab:results_patent_school}
\end{table*}

\section{Experiments}

We conducted a series of experiments to evaluate the performance of our proposed model on multiple temporal graph datasets. These experiments aimed to assess the model's ability to handle dynamic graph data, evaluate its clustering quality, and compare it against existing state-of-the-art methods.

\subsection{Datasets}

We selected several publicly available datasets to test our model, including DBLP, Brain, Patent, and School datasets. Each dataset contains temporal graph information with nodes and edges. Below is a brief description of each dataset:

\begin{itemize}
    \item DBLP \cite{zuo2018embedding}. A co-authorship network from the DBLP computer science bibliography. Nodes represent authors, and edges represent collaborations between them over time. This dataset is commonly used to evaluate clustering algorithms due to its temporal nature and well-defined communities.
    \item Brain \cite{preti2017dynamic}. A dataset representing human brain tissue connectivity. Nodes correspond to specific regions of brain tissue, and edges represent connectivity between these regions over time. This dataset is particularly useful for evaluating the model's performance on biological networks with dynamic interactions.
    \item Patent \cite{hall2001nber}. A citation network of U.S. patents. Nodes represent individual patents, and edges represent citations made by one patent to another. The temporal aspect is defined by the dates of these citations, making this dataset suitable for studying the evolution of innovation and knowledge spread.
    \item School \cite{mastrandrea2015contact}. A dataset recording interactions and friendships between students in a high school. Nodes represent students, and edges represent interactions, which are recorded over different time intervals. This dataset features high-frequency interactions, making it ideal for testing models on small-scale yet dynamic graph structures.
\end{itemize}

In a federated setting, the graph data is partitioned among multiple clients to ensure privacy and distributed learning. Below is a brief description of random partitioning.

In random partitioning, the nodes or edges are randomly assigned to different clients without considering the graph structure. This method is easy to implement but may result in a loss of important structural information, potentially affecting model performance. Random partitioning helps distribute data quickly and easily but might not preserve key relationships within the graph, which could impact learning outcomes.

\subsection{Evaluation Metrics}

To evaluate the clustering performance of the model, we used commonly used evaluation metrics, including clustering accuracy (ACC), normalized mutual information (NMI), adjusted Rand index (ARI), and F1 score (F1). These metrics were used to assess the quality of the clustering and compare it with the results of other methods.

\subsection{Baselines}

We chose several classical and state-of-the-art temporal graph clustering methods as benchmarks, including but not limited to DeepWalk \cite{perozzi2014deepwalk}, GAE \cite{kipf2016variational}, HTNE \cite{zuo2018embedding}, TGAT \cite{xu2020inductive}, JODIE \cite{kumar2019predicting}, TGN \cite{rossi2020temporal}, TREND \cite{wen2022trend}, and TGC \cite{liu2023deep}. The comparison experiments showed that our proposed method achieved better clustering results on multiple datasets compared to existing methods.

\subsection{Results Analysis}

\Cref{tab:1} and \Cref{tab:2} are includes the FTGC performance with different number of clients. The experiments on the DBLP and Brain datasets evaluate the impact of varying the number of clients (k=10, 15, 20, 25) on the performance of the FTGC model. As the number of clients increases, the accuracy (ACC), normalized mutual information (NMI), adjusted Rand index (ARI), and F1 score consistently improve. This indicates that increasing the number of clients enhances the model's collaboration capability and generalization performance. Specifically, on the DBLP dataset, the metrics reach their highest values when the number of clients is 25, reflecting better clustering quality.

\Cref{tab:results_dblp_brain} and \Cref{tab:results_patent_school} are the comparison of methods on different datasets. The experimental results on the DBLP, Brain, Patent, and School datasets compare FTGC-20 (our model) with other state-of-the-art methods. On the DBLP and Brain datasets, FTGC-20 outperforms other methods in all evaluation metrics (ACC, NMI, ARI, and F1). The significant improvement in NMI and F1 scores, in particular, demonstrates that the model can effectively capture temporal information in dynamic graphs. Similarly, on the Patent and School datasets, FTGC-20 achieves the best results, with the accuracy reaching 99.8

Overall, these results demonstrate that the FTGC model, with its temporal aggregation mechanism and federated optimization strategy, enhances clustering performance in dynamic graphs while preserving data privacy. Compared to traditional centralized models, FTGC shows competitive results across multiple datasets, particularly when the number of clients is higher, validating the effectiveness of federated learning in dynamic graph scenarios.

\section{Conclusion}

In this paper, we have presented a novel framework for Federated Temporal Graph Clustering (FTGC) that combines the benefits of federated learning and temporal graph analysis to effectively perform clustering in dynamic graphs while preserving data privacy. Our proposed temporal aggregation mechanism captures both spatial and temporal dependencies, enhancing the quality of learned representations. The federated optimization strategy ensures efficient collaboration among clients without compromising sensitive data, and the integration of model compression techniques significantly reduces communication overhead. Extensive experiments on various temporal graph datasets have demonstrated the effectiveness of our approach, achieving competitive clustering results compared to centralized methods. The proposed FTGC framework offers a promising solution for real-world applications involving sensitive and dynamic data, and paves the way for future research into more efficient and robust privacy-preserving temporal graph analysis techniques.

\bibliographystyle{named}
\bibliography{ijcai23}

\end{document}